\def\year{2019}\relax
\documentclass[letterpaper]{article}
\usepackage{aaai19}
\usepackage{times}
\usepackage{helvet}
\usepackage{courier}
\usepackage{url}
\usepackage{graphicx}
\usepackage{subcaption}
\usepackage{mwe}
\usepackage{algorithm}
\usepackage[noend]{algpseudocode}
\usepackage{multirow}

\usepackage{xcolor,colortbl}
\definecolor{mygreen}{rgb}{0.82421875,0.82421875,0.82421875}
\newcolumntype{b}{>{\columncolor{mygreen}}c}

\title{A Meta-Learning Approach for Custom Model Training}
 \author{Amir Erfan Eshratifar,\textsuperscript{1}
         Mohammad Saeed Abrishami,\textsuperscript{1}
         David Eigen,\textsuperscript{2}
         Massoud Pedram,\textsuperscript{1}\\
\textsuperscript{1}{Department of Electrical Engineering at University of Southern California, Los Angeles, CA 90089, USA}\\
\textsuperscript{2}{Clarifai, San Francisco, CA 94105, USA}\\
eshratif@usc.edu,
abri442@usc.edu, 
deigen@clarifai.com,
pedram@usc.edu}

\begin{document}
\maketitle
\begin{abstract}
\begin{quote}

Transfer-learning and meta-learning are two effective methods to apply knowledge learned from large data sources to new tasks. In few-class, few-shot target task settings (i.e. when there are only a few classes and training examples available in the target task), meta-learning approaches that optimize for future task learning have outperformed the typical transfer approach of initializing model weights from a pre-trained starting point. But as we experimentally show, meta-learning algorithms that work well in the few-class setting do not generalize well in many-shot and many-class cases. In this paper, we propose a joint training approach that combines both transfer-learning and meta-learning. Benefiting from the advantages of each, our method obtains improved generalization performance on unseen target tasks in both few- and many-class and few- and many-shot scenarios.

\end{quote}
\end{abstract}
\begin{table*}[ht]
\caption{Accuracy results on miniImageNet dataset.}
  \centering
\label{tab:miniimagenet_results}
\begin{tabular}{c|c|c|c|c|b|c|b}
\multicolumn{2}{c|}{Task}   & Transfer Learning & Prototypical & Reptile & Reptile MTL & MAML & MAML MTL\\
\hline
\multirow{3}{*}{5-ways}  & 1 Shot   &   37.44    & 49.42   & 49.16    & \textbf{51.04}  &   48.70    &  \textbf{50.99}\\
                         & 5 Shots  &   53.28    & 68.20   & 65.99    & \textbf{69.58}  &   63.11   &   \textbf{67.88}\\
                         & 100 Shots &   90.23    & 61.13   & 83.75    & \textbf{96.56}  &   82.53    & \textbf{92.44}\\
\hline
\multirow{3}{*}{20-ways} & 1 Shot   &   15.06    & 21.54   & 20.29    & \textbf{22.27}  &   20.50    & \textbf{22.34}\\
                         & 5 Shots  &   27.33    & 34.68   & 31.46    & \textbf{36.45}  &   31.50    & \textbf{35.95}\\
                         & 100 Shots &   73.56    & 56.76   & 68.59    & \textbf{74.00}  &   68.55    & \textbf{74.87}\\
\hline
\multirow{3}{*}{35-ways} & 1 Shot   &   10.49    & 10.67   & 9.85    & \textbf{13.60}  &   9.61    & \textbf{14.11}\\
                         & 5 Shots  &   20.04    & 17.53   & 16.46    & \textbf{21.59}  &   16.01    & \textbf{21.85} \\
                         & 100 Shots &   61.72    & 38.13   & 51.09    & \textbf{68.10}  &   50.21    & \textbf{66.34}
\end{tabular}
\end{table*}
\section{Introduction}
Current deep learning algorithms require a very large amount of data to learn decent task-specific models, and acquiring enough labeled data is often expensive and laborious. Moreover, in many mission critical applications, such as autonomous vehicles and drones, an agent needs to adapt rapidly to unseen environments. Humans are able to learn new skills and concepts rapidly by leveraging knowledge learned earlier; therefore, we aim to enable the artificial agents to do the same. Transfer learning transfers the knowledge obtained from one domain with a large amount of labeled data to other domains with less labeled data~\cite{Pan}. It achieves this by copying the initial feature extraction layers, and fine-tuning the resulting model on the target task~\cite{Yosinski}. However, this method is still data hungry because gradient-based optimization algorithms need many iterations over numerous examples to adapt the models for new tasks~\cite{Ravi}. On the other hand, meta-learning is a class of machine learning algorithms concerned with the ability of learning process itself. Introduced by \cite{Schmidhuber}, meta-learning aims to train the model in task space rather than instance space. While transfer learning methods train a base model to use as a transfer source by optimizing a single monolithic task, meta-learning algorithms learn their base models by sampling many different smaller tasks from a large data source. As a result, one might expect that the meta-learned model is capable of generalizing well to new unseen tasks because of task-agnostic way of training.

\textbf{Shortcomings of meta-learning algorithms.} As we show in the experiments below, models trained using meta-learning perform worse than transfer learning in the following two scenarios: \textbf{1.} When there are many training examples available for each class in the target task (here we would like the artificial agent to continue improving its model performance as more data becomes available); and \textbf{2.} When there are many different classes in the target task.

The main contribution of this paper is a joint ``meta-transfer'' learning method that performs well for target tasks of both few and many shots/classes. Our method performs better than both transfer- and meta-learning baselines on all target task sizes we evaluate.


\section{Meta-Transfer Learning (MTL)}

In order to overcome the two issues mentioned earlier, we propose a new training algorithm, which inherits advantages of both meta-learning and transfer learning. This joint training method employs two loss functions: 1) task-specific (transfer learning) 2) task-agnostic (meta-learning). The task-specific loss, $L_{(x,y)}(\theta)$, is defined over the entire base model's training dataset. The task-agnostic loss, $L_{\tau}(\theta)$, on the other hand, is a meta-learning loss defined over a distribution of tasks (e.g. 5-ways classification tasks). Two gradient updates are computed independently from these two loss functions, and the model is updated using the weighted average of these two update vectors (see Algorithm 1). The tasks in meta-learning are sampled from a distribution $p(\tau)$, while all instances in the sampled tasks are used for the task-specific optimization. For adaptation to a new unseen task, regular stochastic gradient descent will be used.
For the meta-learner, we evaluate our method using both Model Agnostic Meta-learning (MAML~\cite{MAML}) and its first order variant, Reptile~\cite{Reptile}. The reason that we use this class of meta-learning algorithms is that as opposed to Matching Networks~\cite{MatchingNetwork} and its variant~\cite{Prototypical}, they are model agnostic, and can be directly applied to any model which is trained with a gradient descent procedure. The proposed method is similar to Gradient Agreement~\cite{GradientAgreement} in a sense that it pushes the model parameters in a direction that the distribution of tasks have agreement with the single specific task of training over the whole classes.

\begin{figure}[H]   
  \begin{center}
    \includegraphics[width=.95\columnwidth]{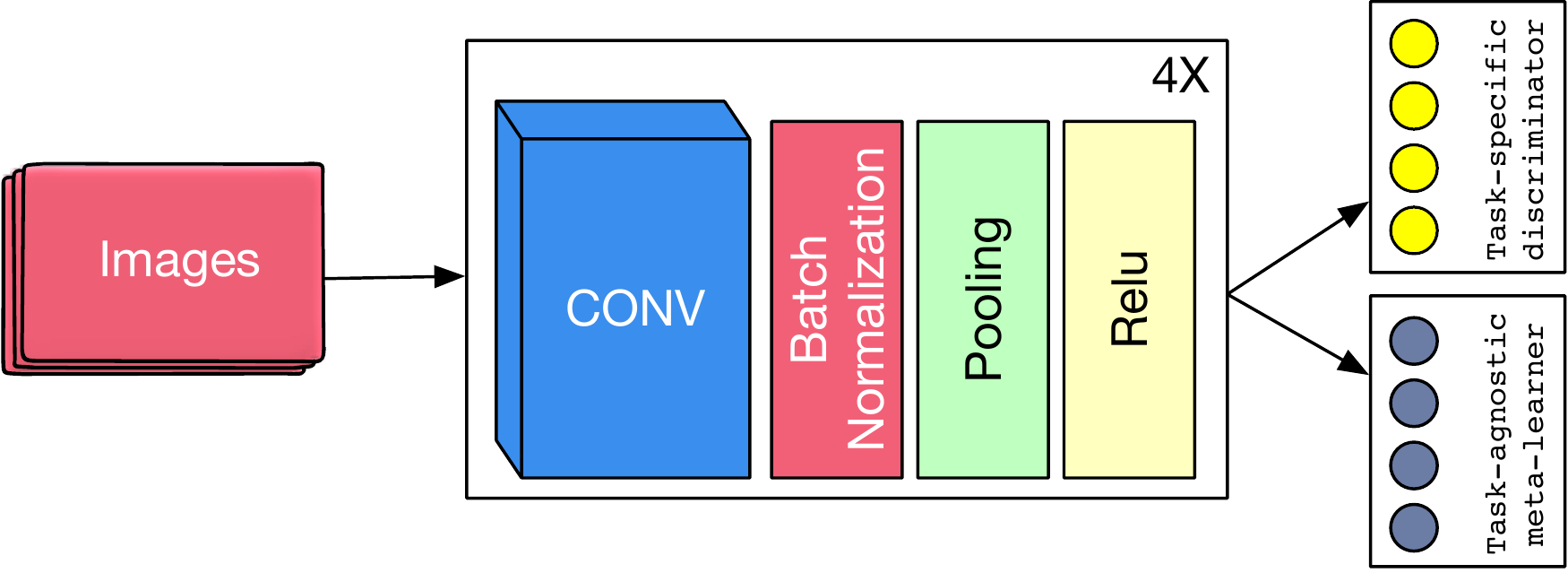}
    \caption{Meta-transfer learning model setup for miniImageNet dataset}\label{meta-transfer_model}
  \end{center}
\end{figure}

\begin{algorithm}
\caption{Meta-Transfer Learning Algorithm}\label{basic}
\begin{algorithmic}[1]  
	\State {Initialize model parameters, $\theta$}
     \For{\texttt{iteration = 1,2,...}}
        \State Sample a batch of tasks $\tau_i$ $\sim$ p($\tau$)
        \For{all $\tau_i$}
        \State Split the examples of the task into $k$ sub-batches
        \State $\theta_i$ = $\theta$ - $\alpha_{inner} \nabla L_{\tau_i}(f_\theta)$ for $k$ steps
        \EndFor
        \State $\theta_{metalearner}$ = $\theta$ - $\alpha_{outer}$ $\sum_{i}$ $\nabla L_{\tau_i}(f_{\theta_i})$ (MAML)
        \State $\theta_{metalearner}$ = $\theta$ + $\alpha_{outer}$ $\sum_{i}$($\theta_i$ - $\theta$) (Reptile)
        \State $\theta_{discriminator}$ = $\theta$ - $\alpha_{d} \nabla L_{(x,y)}$
        \State $\theta$ = $\beta~\theta_{metalearner}$ + $(1-\beta)~\theta_{discriminator}$
      \EndFor
	\State \Return $\theta$
\end{algorithmic}
\end{algorithm}

\section{Performance Evaluations}
The proposed model is evaluated on \textit{miniImageNet}~\cite{MatchingNetwork} dataset, split into 64 training classes and 36 test classes as unseen tasks. The architecture of the model is shown in Figure~\ref{meta-transfer_model} and the results are demonstrated in Table~\ref{tab:miniimagenet_results}. The base model for transfer learning is trained on all 64 training classes.

Note that for many-classes (35-ways) tasks, the transfer learning baseline outperforms previous meta-learning algorithms, while in few-classes problems, the result is reversed: meta-learning beats transfer learning. Our proposed method, MTL, outperforms both these algorithms in all scenarios by improving the weaknesses of few-shot learning algorithms in generalizing to many-shot and many-classes problems. 

\section{Conclusion and Future Work}
A single model that is adaptable to unseen tasks is a crucial component in artificial intelligence. In this work, we presented a method to extend the capability of few-shot learning algorithms to many-shot and many-classes learning problems, by integrating them with transfer learning model. The next step is to use this approach on a larger dataset and deeper model, to see whether meta-learning is still outperforming transfer learning or not.

\bibliographystyle{aaai}
\bibliography{references.bib}

\begin{thebibliography}{}

\bibitem[\protect\citeauthoryear{Eshratifar \bgroup et al\mbox.\egroup
  }{2018}]{GradientAgreement}
Eshratifar, A.~E.; Eigen, D.; and Pedram, M.
\newblock 2018.
\newblock Gradient agreement as an optimization objective for meta-learning.
\newblock {\em CoRR} abs/1810.08178.

\bibitem[\protect\citeauthoryear{Finn \bgroup et al\mbox.\egroup }{2017}]{MAML}
Finn, C.; Abbeel, P.; and Levine, S.
\newblock 2017.
\newblock Model-agnostic meta-learning for fast adaptation of deep networks.
\newblock {\em CoRR} abs/1703.03400.

\bibitem[\protect\citeauthoryear{Nichol \bgroup et al\mbox.\egroup
  }{2018}]{Reptile}
Nichol, A.; Achiam, J.; and Schulman, J.
\newblock 2018.
\newblock On first-order meta-learning algorithms.
\newblock {\em CoRR} abs/1803.02999.

\bibitem[\protect\citeauthoryear{Pan \bgroup et al\mbox.\egroup }{2010}]{Pan}
Pan, S.~J., and Yang, Q.
\newblock 2010.
\newblock A survey on transfer learning.
\newblock {\em IEEE Transactions on Knowledge and Data Engineering}
  22(10):1345--1359.

\bibitem[\protect\citeauthoryear{Ravi \bgroup et al\mbox.\egroup }{2016}]{Ravi}
Ravi, S., and Larochelle, H.
\newblock 2016.
\newblock Optimization as a model for few-shot learning.
\newblock {\em ICLR}.

\bibitem[\protect\citeauthoryear{Schmidhuber}{1987}]{Schmidhuber}
Schmidhuber, J.
\newblock 1987.
\newblock Evolutionary principles in self-referential learning. on learning now
  to learn: The meta-meta-meta...-hook.
\newblock Diploma thesis, Technische Universitat Munchen, Germany.

\bibitem[\protect\citeauthoryear{Snell \bgroup et al\mbox.\egroup
  }{2017}]{Prototypical}
Snell, J.; Swersky, K.; and Zemel, R.~S.
\newblock 2017.
\newblock Prototypical networks for few-shot learning.
\newblock {\em CoRR} abs/1703.05175.

\bibitem[\protect\citeauthoryear{Vinyals \bgroup et al\mbox.\egroup
  }{2016}]{MatchingNetwork}
Vinyals, O.; Blundell, C.; Lillicrap, T.~P.; Kavukcuoglu, K.; and Wierstra, D.
\newblock 2016.
\newblock Matching networks for one shot learning.
\newblock {\em CoRR} abs/1606.04080.

\bibitem[\protect\citeauthoryear{Yosinski \bgroup et al\mbox.\egroup
  }{2014}]{Yosinski}
Yosinski, J.; Clune, J.; Bengio, Y.; and Lipson, H.
\newblock 2014.
\newblock How transferable are features in deep neural networks?
\newblock {\em CoRR} abs/1411.1792.

\end{thebibliography}

\end{document}